\documentclass{article}
\usepackage{ICASSP}
\usepackage{hyperref}
\usepackage{graphicx}
\usepackage{booktabs}
\usepackage{amssymb,amsmath,bm}
\usepackage{textcomp}
\usepackage{graphicx}
\usepackage[textfont=it,tableposition=top]{caption}

\title{multilevel Transformer FOR MULTIMODAL EMOTION RECOGNITION}
\name{Junyi He \qquad Meimei Wu \qquad Meng Li \qquad Xiaobo Zhu\qquad Feng Ye\qquad }
\address{360 DigiTech, China}

\begin{document}

\maketitle
\begin{abstract}
Multimodal emotion recognition has attracted much attention recently. Fusing multiple modalities effectively with limited labeled data is a challenging task. Considering the success of pre-trained model and fine-grained nature of emotion expression, we think it is reasonable to take these two aspects into consideration. Unlike previous methods that mainly focus on one aspect, we introduce a novel multi-granularity framework, which combines fine-grained representation with pre-trained utterance-level representation. Inspired by Transformer TTS, we propose a multilevel transformer model to perform fine-grained multimodal emotion recognition. Specifically, we explore different methods to incorporate phoneme-level embedding with word-level embedding. To perform multi-granularity learning, we simply combine multilevel transformer model with Bert. Extensive experimental results show that multilevel transformer model outperforms previous state-of-the-art approaches on IEMOCAP dataset. Multi-granularity model achieves additional performance improvement.
\end{abstract}

\section{Introduction}

Speech emotion recognition (SER) is a promising area of research mainly for human-computer interaction system, which aims to recognize the emotion state (such as happy, angry, and sad) of a speaker from his/her speech \cite{2007Emotion}. There are mainly two challenges for the SER task. One is the lack of large-scale labeled data since labeling emotion is subjective and multi-person annotation is needed, which requires a lot of time and human effort \cite{DBLP:conf/interspeech/ZhaoWW22}. The other challenge is that emotion expression is multimodal and fine-grained \cite{DBLP:conf/interspeech/LiDWL21}. How to combine different modalities effectively is a long way to explore.

One common solution to limited labeled data is to leverage transfer learning-based approaches. Recently a class of techniques known as self-supervised learning (SSL) architectures have achieved state-of-the-art (SOTA) performance in natural language processing (NLP) \cite{DBLP:conf/naacl/DevlinCLT19,DBLP:conf/iclr/LanCGGSS20,DBLP:journals/corr/abs-1907-11692} and speech recognition \cite{DBLP:conf/interspeech/SchneiderBCA19,DBLP:conf/nips/BaevskiZMA20,DBLP:conf/iclr/BaevskiSA20}. For emotion recognition task, F. A. Acheampong et al. \cite{DBLP:journals/air/AcheampongNC21} and L. Pepino et al. \cite{DBLP:conf/interspeech/PepinoRF21} have done great jobs with text or speech pre-trained models respectively. However the above methods only focused on one modality.

A large number of approaches have been developed to learn the interaction between different modalities. For the approaches with the pre-trained models, S. Siriwardhana et al. \cite{DBLP:conf/interspeech/SiriwardhanaRWN20} and Z. Zhao et al. \cite{DBLP:conf/interspeech/ZhaoWW22} explored early fusion and late fusion of text and speech representations respectively for emotion recognition leveraging both Bert \cite{DBLP:conf/naacl/DevlinCLT19} related and Wav2vec \cite{DBLP:conf/interspeech/SchneiderBCA19} related models. The results showed that late fusion models generally got better results. However, the above interaction of late fusion between different modalities was only based on aggregated pre-trained text and speech embedding. For the approaches without pre-trained models, researchers utilized different models \cite{DBLP:conf/slt/0002BJ18,DBLP:conf/interspeech/XuZHWPL19,DBLP:conf/interspeech/LiDWL21} to interact different modalities. S. Yoon et al. \cite{DBLP:conf/slt/0002BJ18} built a deep neural network with recurrent neural networks (RNN) to learn vocal representations and text representations and then concatenated them for emotion classification. However the approach was based on utterance-level fusion. H. Xu et al. \cite{DBLP:conf/interspeech/XuZHWPL19} proposed a fine-grained method to learn the alignment between speech and text, together with long short-term memory (LSTM) network to model the sequence for emotion recognition. Nevertheless, LSTM only consumed the input sequentially and the interaction within a single modality was not fully explored. H. Li et al. \cite{DBLP:conf/interspeech/LiDWL21} proposed a fine-grained emotion recognition model with a temporal alignment mean-max pooling operation and cross-modality mechanism. Nevertheless, it required additional work for labeling and alignment prediction to use aligned information in production system.

SSL enables us to use a large unlabelled dataset to train models that can be later used to extract representations and fine-tune for specific problems with limited amount of labeled data \cite{DBLP:conf/interspeech/SiriwardhanaRWN20}. However, the above-mentioned aggregated pre-trained embedding is a good representation for the entire sentence, not for the specific words or voice fragments. To further improve the performance of SER, we need to explore an effective way to add fine-grained interaction between different modalities with limited additional human effort. Transformer TTS \cite{DBLP:conf/aaai/Li0LZL19} is a fine-grained model in the text to speech (TTS) area. For the training process, with the phoneme and mel sequence as input, Transformer TTS network generates mel spectrogram. Inspired by this work, we can use the similar structure to utilize audio and text information at fine-grained level without additional alignment. Phoneme sequence plays an important role to generate mel spectrogram in the TTS task. For the SER task, sometimes the stress of a sentence is on some specific phonemes, however compared with the word input, only phoneme information is not enough.

To overcome above challenges, we propose a novel multi-granularity framework to merge pre-trained utterance-level representation with fine-grained representation. For the fine-grained part, we propose a multilevel transformer model to introduce cross-modal interaction among voice fragments, words, and phonemes. We compare different methods to incorporate the phoneme embedding with word embedding. Vanilla transformer \cite{DBLP:conf/nips/VaswaniSPUJGKP17} is added to further aggregate the sequential multimodal representations. To perform multi-granularity learning, we simply combine multilevel transformer model with the pre-trained model. In this article, the pre-trained model that we choose is Bert \cite{DBLP:conf/naacl/DevlinCLT19}. Our experimental results on the Interactive Emotional Dyadic Motion Capture (IEMOCAP) \cite{DBLP:journals/lre/BussoBLKMKCLN08} dataset show that multilevel transformer model achieves state-of-the-art results. The multi-granularity model yields additional performance boost.

In summary, our main contributions are as follows:

• We conduct fine-grained learning with multilevel transformer model (Section 3.1) to obtain fine-grained cross modality information from voice fragments, words, and phonemes.

• We propose a simple but effective multi-granularity fusion framework to combine fine-grained representation with pre-trained utterance-level representation (Section 3.2).

• We design and evaluate our approaches quantitatively on IEMOCAP dataset. Experimental results show that multilevel transformer model outperforms existing state-of-the-art methods. Multi-granularity model gives additional performance improvement (Section 4).

\section{Related work}

After the classical machine learning models such as the Hidden Markov Model \cite{DBLP:journals/speech/NweFS03} and the Gaussian Mixture Model \cite{DBLP:conf/icassp/AyadiKK07}, were employed based on handcrafted low-level features or statistical high-level features, models with deep neural networks have been actively studied in SER. D. Bertero et al. \cite{DBLP:conf/icassp/BerteroF17} proposed the model consisting of the convolution neural network (CNN) that extracted high-level features from raw spectrogram features. A. Satt et al. \cite{DBLP:conf/interspeech/SattRH17} proposed an end-to-end model with CNN and LSTM network to capture the contextual information.

Recently, multimodal models that make use of both audio and text information for SER have attracted much attention. S. Yoon et al. \cite{DBLP:conf/slt/0002BJ18} employed RNN to encode audio and text and then used the last hidden state of a recurrent modality encoder as a query and used the other encoded modality as a key-value pair in the attention mechanism. However the interaction between different modalities was not fully explored. H. Xu et al. \cite{DBLP:conf/interspeech/XuZHWPL19} designed the model with LSTM to learn the alignment between the audio and text from the attention mechanism. However, the interaction within a single modality was not fully explored. H. Li et al. \cite{DBLP:conf/interspeech/LiDWL21} proposed a fine-grained emotion recognition model from aligned audio and text by using temporal mean-max alignment pooling method and cross modality module, which needed aligned audio and text as input. 

\section{Proposed Methods}

In this section, we first introduce our multilevel transformer model. Then, we present our multi-granularity model, consisting of multilevel transformer model and Bert.

\begin{figure}[t]
  \centering
  \includegraphics[scale=0.55]{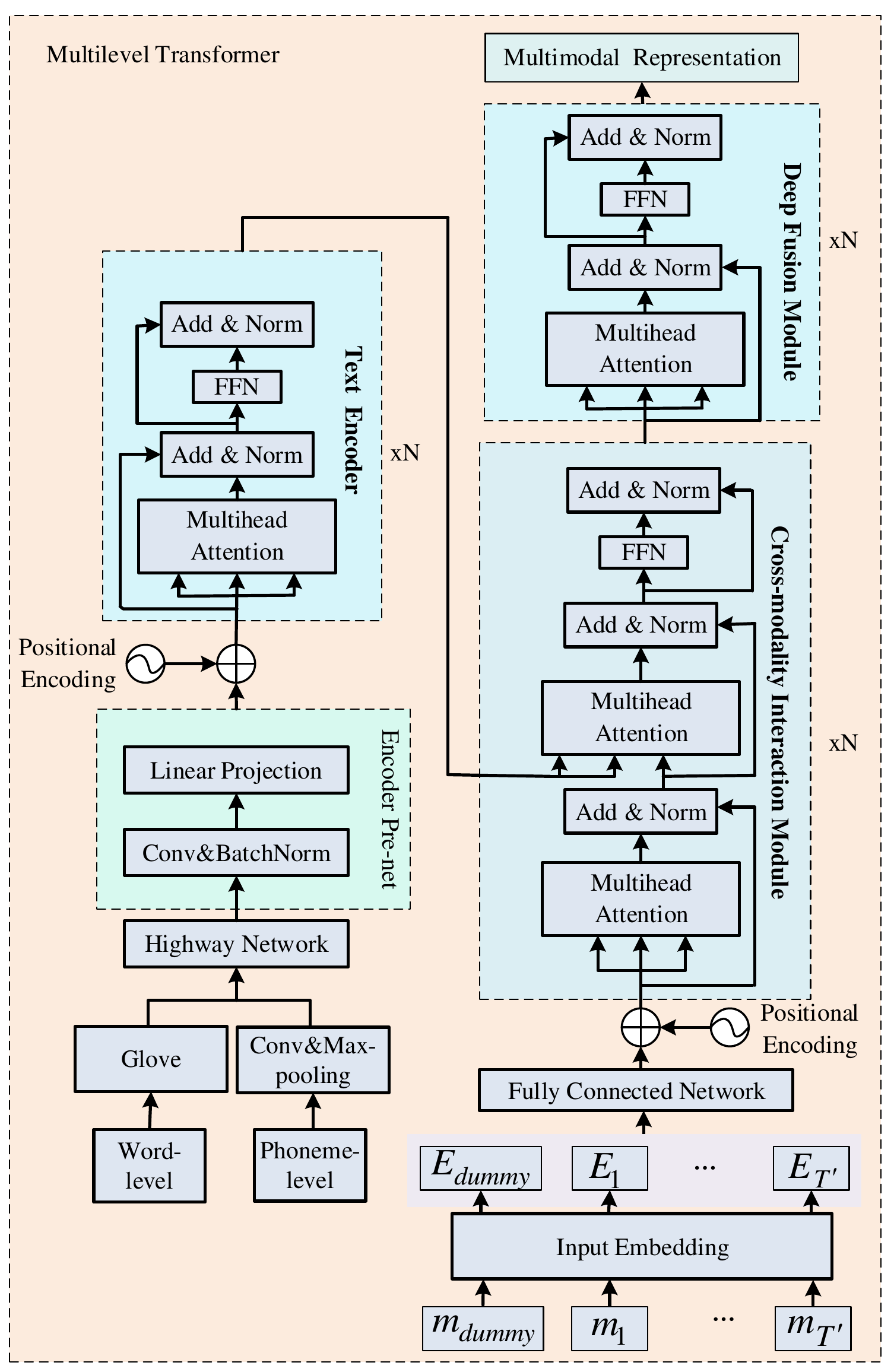}
  \caption{Overview of multilevel transformer.}
  \label{fig:speech_production1}
\end{figure}

\subsection{Multilevel transformer model}

Here we first introduce the overall architecture of our multilevel transformer model. Then we focus on the detailed parts.

\subsubsection{Architecture}

Transformer TTS \cite{DBLP:conf/aaai/Li0LZL19} is a neural TTS model based on Tacotron2 \cite{DBLP:conf/icassp/ShenPWSJYCZWRSA18} and transformer \cite{DBLP:conf/nips/VaswaniSPUJGKP17}. Inspired by Transformer TTS, we propose our multilevel transformer model. As shown in Fig 1, text input is firstly transformed into phoneme and word. After that, the output is processed by the highway network \cite{DBLP:journals/corr/SrivastavaGS15}, followed by encoder pre-net (3-layer CNN and 1-layer projection), and then is fed into the text encoder. The mel spectrogram is processed with a 2-layer fully connected network. The output of fully connected network and previous text encoder is sent into cross-modality interaction module, followed by deep fusion module. The output of the deep fusion module is used to predict emotion category probability.

\subsubsection{Overall process}

For the SER task, we utilize the actual mel spectrogram and text information not only in the training stage, but also in the inference stage. What's more, similar to BERT’s [class] token, we prepend one dummy mel vector to the sequence of actual mel spectrogram  $m=(m_{dummy},m_1,m_2,...,{m_{T^{'}}})$. In TTS task, the dummy mel vector is used to predict the first mel spectrogram. In our scenario, it is used to the calculate the final aggregated representation.

For the TTS task, during the inference stage, TTS converts an input text sequence $x=(x_1,x_2,...,x_T)$ into an output mel spectrogram sequence $o=(o_1,o_2,...,{o_{T^{'}}})$ and each predicted $o_t$ is conditioned on predicted outputs $o_1,o_2,...,o_{t-1}$. This conversion can be formulated as the following conditional probability:

\begin{equation}
  f(o_t|x_1,...,x_{T}) =  f(o_t|o_{<t},x)
\end{equation}

Nevertheless, our task is to predict the emotion category with whole available information, not to predict mel spectrogram in a sequence to sequence manner. In the inference stage, we combine the text input with whole golden mel spectrogram sequence instead of the previous predicted one. Thus the emotion category probability $p$ can be computed by:

\begin{equation}
 p = g(x,m)
\end{equation}
where $g(x,m)$ is the function that calculates the probability of each emotion category with text input and mel spectrogram.

\subsubsection{Text to phoneme and word embedding}

Similar to Transformer TTS, text input is firstly transformed into the phoneme sequence, which carries fine-grained information. In our scenario, phoneme information is also useful since in some circumstance, the focus of the emotion is on some specific phonemes.

Following \cite{DBLP:conf/iclr/SeoKFH17,DBLP:conf/emnlp/Kim14}, we obtain the phoneme level embedding of each word using CNN. The outputs of the CNN are max-pooled over the entire width to obtain a fixed-size vector for each word. 

In addition, the word level information is also important for emotion recognition. So we add the word embedding via Glove \cite{DBLP:conf/emnlp/PenningtonSM14} since it carries additional fine-grained information.

\subsubsection{Combination of phoneme and word embedding}
Here we explore two different combination methods.

Concatenation is a straightforward way to combine phoneme embedding with word embedding. For the first method, the above embedding is simply concatenated, followed by encoder pre-net.

For the other method, we try to use highway network \cite{DBLP:journals/corr/SrivastavaGS15} since it usually utilizes the gating mechanism to pass information efficiently through several layers. With the reference of \cite{DBLP:conf/iclr/SeoKFH17}, The concatenation of the phoneme and word embedding vectors $u$ is passed to a two-layer highway network to fuse multi-level information effectively:

\begin{equation}
  Z(u) = H(u) \cdot T(u) + u \cdot (1 - T(u))
\end{equation}
where $H(u)$ is a parametric transformation (an affine projection followed by ReLU \cite{DBLP:journals/jmlr/GlorotBB11}) of the input $u$ and $T(u)$ is a gating unit, which controls how much transformation is applied and how much copy of the original input is activated. Then, the multi-level textual information is processed by the encoder pre-net to model long-term context.

\subsubsection{Transformer modules}

We use vanilla transformer \cite{DBLP:conf/nips/VaswaniSPUJGKP17} structure for text encoder, cross-modality interaction module and deep fusion module. Text encoder contains self-attention layer to fuse information from words and phonemes. Cross-modality interaction module includes self-attention layer and encoder-decoder attention layer to integrate the output from text encoder with phoneme information. We add vanilla transformer blocks including self-attention layer to deeply fuse multimodal sequential representations after the cross-modality interaction module. Finally, corresponding to the dummy mel input, we take the first output vector of deep fusion module as the global representation and apply a linear projection based on it with logits output.

\subsubsection{Loss}

For the TTS task, Transformer TTS model generates mel spectrogram and stop token. The predicted ones are compared with the ground truth to calculate the TTS loss.

Inspired by \cite{DBLP:conf/interspeech/CaiYZ0021}, we try to adopt multi-task learning to optimize the joint loss of TTS and SER, however the performance does not improve in our scenario. So only the logits of the last projection layer are used to classify the input example, with cross entropy as the loss function:

\begin{equation}
  L = -\sum_{i=1}^{N}\sum_{k=1}^{K} y_{i,k} log p_{i,k}
\end{equation}
where $p_{i,k}$ is the probabilistic value that indicates the final probabilities of class $k$ of utterance $i$ and $y_{i,k}=1$ if the ${i}$-th sample belongs to ${k}$-th class.

\subsection{Multi-granularity model}

\subsubsection{Basic components introduction}

Bert has achieved SOTA performance in natural language processing. Bert model consists of 12 layers and the embedding dimension size is 768.

For the multilevel transformer part, the structure is the same with our previous introduction.

\begin{figure}[t]
  \centering
  \includegraphics[scale=0.5]{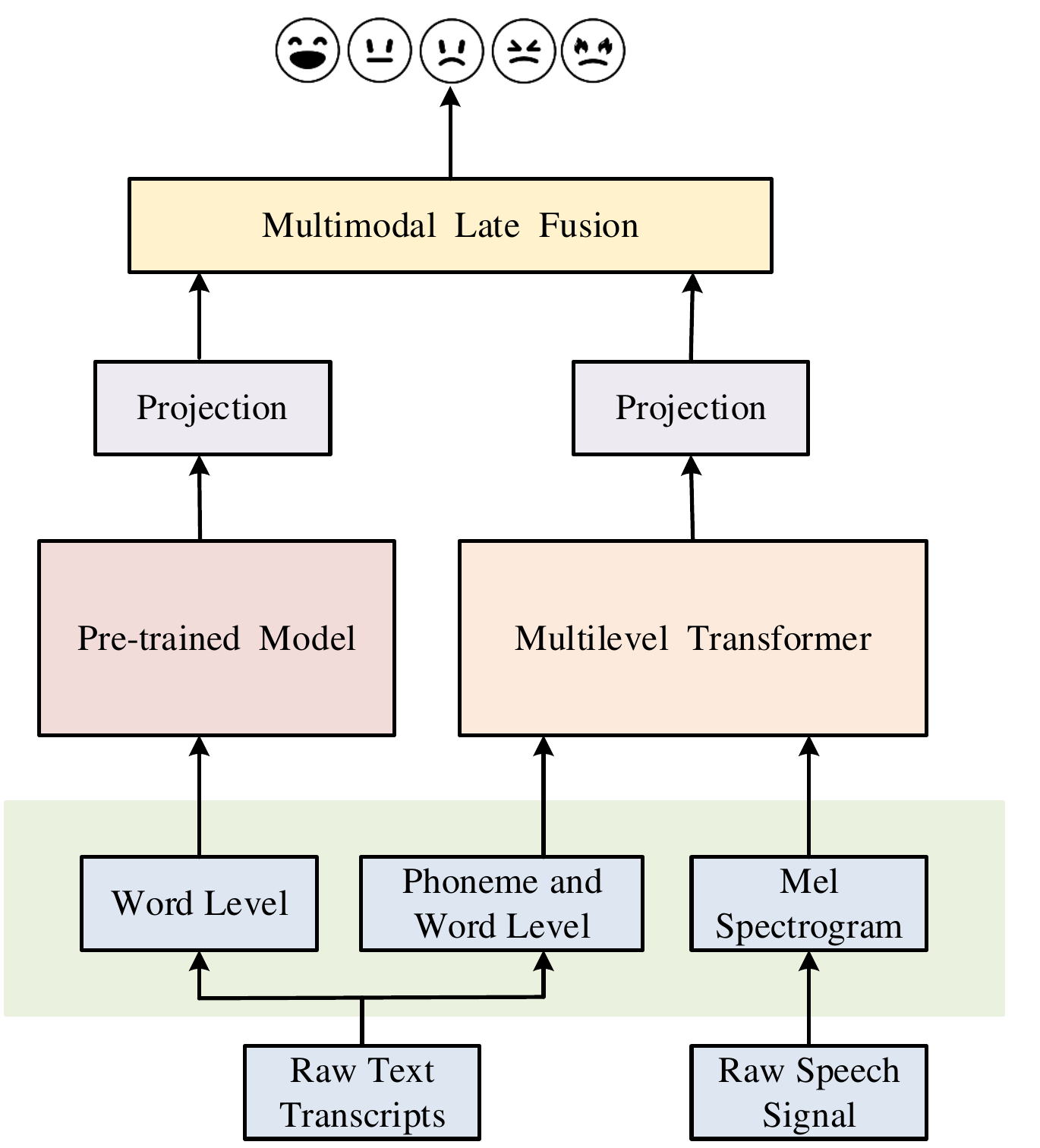}
  \caption{Overview of the multi-granularity model.}
  \label{fig:speech_production2}
\end{figure}

\subsubsection{Model pipeline}

 Our multi-granularity model is depicted in Fig.2. The success of Bert model in sentence classification tasks highlights the effective use of the CLS token, which can be used as a representation for the entire sequence \cite{DBLP:conf/interspeech/SiriwardhanaRWN20}. Hence, we utilize the CLS embedding of Bert to obtain pre-trained utterance representation. 
 
 The CLS embedding generated from deep fusion module of multilevel transformer model is used to provide fine-grained multimodal representation.

A late fusion mechanism followed by a classification head works remarkably well with fine-tuned “Bert-Like” pre-trained SSL models even in a multimodal setting \cite{DBLP:conf/interspeech/SiriwardhanaRWN20}. With the reference of \cite{DBLP:conf/naacl/DaiCLF21}, after the above CLS embedding of different models is processed by the projection respectively, we simply concatenate the outputs. Finally we send the concatenated embedding through the classification head, which includes a fully connected layer that outputs logits.

\section{Experiments}


\subsection{Dataset}
We use the IEMOCAP \cite{DBLP:journals/lre/BussoBLKMKCLN08} dataset, which is the most widely used dataset in emotion recognition research. It contains approximately 12 hours of audiovisual data, including video, speech and text transcriptions. We use audio and transcriptions only in this research.

To be comparable with previous related researches \cite{DBLP:conf/slt/0002BJ18}, 4 categories of emotions are used: angry (1103 utterances), sad (1084 utterances), neutral (1708 utterances) and happy (1636 utterances, merged with excited), resulting in a total of 5531 utterances. We perform a 5-fold cross-validation with 3, 1, 1 in train, dev, and test sets respectively. Every experiment is run for 3 times to reduce randomness, and the averaged result is used as the final performance score.


\subsection{Implementation detail}
We implement the proposed models by using the PyTorch deep learning framework. For the acoustic data, we extract the 128-dimensional filterbank features from speech signals. The window size and hop size are set to 25ms and 12ms respectively. For the text data, we use 300-dimensional Glove \cite{DBLP:conf/emnlp/PenningtonSM14} embedding for the word. The hidden size of all transformer layers is set to 128. Both models are trained on a Tesla V100 GPU. Adam optimizer \cite{DBLP:journals/corr/KingmaB14} is chosen. Learning rate is set to 1e-5 and batch size is set to 4. Weighted accuracy (WA) of the validation data set is used as the early stop criteria. Weighted accuracy (WA) and unweighted accuracy (UA) are calculated for test data set.


\subsection{Multilevel transformer performance evaluation}

\begin{table}[t]
  \caption{Comparison between multilevel transformer models and previous state-of-the-art models.}
  \label{tab:example1}
  \centering
  \begin{tabular}{@{}l c c}
    \toprule
    \textbf{Proposed Methods} & \textbf{WA} & \textbf{UA} \\
    \midrule
    S. Yoon et al. \cite{DBLP:conf/slt/0002BJ18}        &{0.690 }$\pm${ 0.011}&{0.696 }$\pm${ 0.013}\\
    H. Xu et al. \cite{DBLP:conf/interspeech/XuZHWPL19} &{0.692 }$\pm${ 0.006}&{0.699 }$\pm${ 0.007}\\
    H. Li et al. \cite{DBLP:conf/interspeech/LiDWL21}   &{0.719 }$\pm${ 0.003}&{0.728 }$\pm${ 0.004}\\
    Our proposal                            &\textbf{0.730 }$\pm$\textbf{ 0.003}&\textbf{0.741 }$\pm$\textbf{ 0.001}\\ 
    \midrule
    \textbf{Ablation Study} &\textbf{WA} &\textbf{UA} \\
    \midrule
    phoneme only              &{0.689 }$\pm${ 0.002}&{0.701 }$\pm${ 0.005}\\
    word only                 &{0.719 }$\pm${ 0.002}&{0.727 }$\pm${ 0.002}\\
    concatenation             &{0.729 }$\pm${ 0.003}&{0.738 }$\pm${ 0.003}\\
    highway network           &\textbf{0.730 }$\pm$\textbf{ 0.003 }&\textbf{0.741 }$\pm$\textbf{ 0.001}\\
    \midrule
    w/o deep fusion module & {0.724 }$\pm${ 0.009}& {0.734 }$\pm${ 0.006}\\
    \bottomrule
  \end{tabular}
 
\end{table}

We evaluate our multilevel transformer model on the IEMOCAP dataset. For a fair comparison, all the approaches are implemented based on the same dataset with 5-fold cross validation configuration. The results are presented in the first block of Table 1. From the table, we can see that multilevel transformer model outperforms all the baseline models. 

For the ablation study, we conduct several experiments to evaluate key factors in our proposed model. In the second block of Table 1, we find that word with Glove \cite{DBLP:conf/emnlp/PenningtonSM14} embedding improves the performance a lot compared with the phoneme sequence. We should also mention that \cite{DBLP:conf/interspeech/LiDWL21} needs the aligned audio and text as input which requires additional work for production use. However, by virtue of the cross-modality interaction module, multilevel transformer model with word input achieves the competitive results without alignment information. The results also show that the input with both word and phoneme information yields additional performance improvement. Compared with the concatenation, highway network lifts the performance slightly. When deep fusion module is replaced with max pooling and dummy mel input is removed, the performance of the model decreases.

We compare the number of transformer layers in different modules to further explore the performance impact. In Table 2, we find that for the IEMOCAP dataset, transformer structure with one or two layers usually gets the better results for multilevel transformer model. Multilevel transformer model achieves the best results with 1-layer text encoder, 1-layer cross-modality interaction module, and 2-layer deep fusion module.

\begin{table}[t]
  \caption{Comparison of the number of different transformer modules for multilevel transformer model.}
  \label{tab:example2}
  \centering
  \begin{tabular}{@{}ccc cc }
    \toprule
    \textbf{Text Encoder}& \textbf{Cross-mod} & \textbf{Deep Fusion}& \textbf{WA} & \textbf{UA} \\
    \midrule
    3&3 &1 &0.725  & 0.736\\
    2&2 &1&0.729 & 0.737\\
    1&1 &1&0.729 & 0.740\\
    1&1 &2&0.730 & \textbf{0.741}\\
    1&1 &3&0.724 & 0.733\\
    2&2 &2&\textbf{0.731} & 0.739\\
    2&2 &3&0.722 & 0.732\\

    \bottomrule
  \end{tabular}

\end{table}



\subsection{Multi-granularity model performance evaluation}


 
\begin{table}[t]
  \caption{Comparison between multilevel transformer model and its components}
  \label{tab:example1}
  \centering
  \begin{tabular}{@{}l c c}
    \toprule
    \textbf{Proposed Methods} & \textbf{WA} & \textbf{UA} \\
    \midrule
    Bert &{0.693 }$\pm${ 0.003}& {0.695 }$\pm${ 0.000}\\
    Multilevel transformer model &{0.730 }$\pm${ 0.003}& {0.741 }$\pm${ 0.001}\\ 
    Multi-granularity model  &\textbf{0.745 }$\pm$\textbf{ 0.003}  &\textbf{0.750 }$\pm$\textbf{ 0.005} \\
    \bottomrule
  \end{tabular}
 
\end{table}

We compare the performance between multi-granularity model and its components. As shown in Table 3, our multi-granularity model shows better results than Bert and multilevel transformer model. It is easy and straightforward to combine fine-grained representation with pre-trained utterance-level representation to further improve the performance for emotion recognition task.

\section{Conclusions}

In this paper, we first propose multilevel transformer model to perform fine-grained interaction between different modalities from voice fragments, words, and phonemes for speech emotion recognition. As per our knowledge, this is the first time that Transformer TTS structure is used in SER task. Then we introduce a multi-granularity framework to integrate fine-grained representation with pre-trained utterance-level representation in a simple but effective way. Extensive experiment results show that multilevel transformer model outperforms existing state-of-the-art methods. Multi-granularity model achieves additional performance improvement. We think this method can be taken as a reference for other pre-trained models. We will make the code publicly available. In future, we will further explore the way to perform multi-granularity emotion recognition with acoustic pre-trained models, such as Wav2vec 2.0.

\bibliographystyle{IEEEtran}

\bibliography{mybib}

\end{document}